\documentclass[sigconf]{acmart}

\AtBeginDocument{%
  }

\copyrightyear{2025}
\acmYear{2025}
\setcopyright{cc}
\setcctype{by}
\acmConference[MM '25]{Proceedings of the 33rd ACM International Conference on Multimedia}{October 27--31, 2025}{Dublin, Ireland}
\acmBooktitle{Proceedings of the 33rd ACM International Conference on Multimedia (MM '25), October 27--31, 2025, Dublin, Ireland}\acmDOI{10.1145/3746027.3754887}
\acmISBN{979-8-4007-2035-2/2025/10}

\usepackage{subcaption}
\usepackage{xspace}
\usepackage{color}
\usepackage{xcolor}

\usepackage{booktabs}
\usepackage{multirow}
\usepackage{graphicx}

\usepackage{pifont} 

\ifodd 1
\newcommand{\com}[1]{\textbf{\color{red}(COMMENT: #1)}} 
\else

\newcommand{\com}[1]{}

\fi
\def\fig{Fig.\xspace}
\def\eq{Eq.\xspace}

\def\tab{Table\xspace}

\def\1figwidth{0.55\textwidth}
\def\2figwidth{0.38\textwidth}
\def\3figwidth{0.31\textwidth}
\def\4figwidth{0.23\textwidth}

\newcommand{\sysname}{{OpenMap}\xspace}
\newcommand{\sysnameposs}{{OpenMap's}\xspace}

\newcommand{\ie}{\emph{i.e.}\xspace}
\newcommand{\eg}{\emph{e.g.}\xspace}

\begin{document}




\title[OpenMap: Instruction Grounding via Open-Vocabulary Visual-Language Mapping]{OpenMap: Instruction Grounding via Open-Vocabulary Visual-Language Mapping}



\author{Danyang Li}
\authornote{Both authors contributed equally to this research.}
\email{lidanyang1919@gmail.com}
\orcid{0000-0002-7527-4669}
\affiliation{%
  \institution{School of Software, Tsinghua University}
  \city{Beijing}
  \country{China}
}
\author{Zenghui Yang}
\authornotemark[1]
\email{zenghuiyang36@gmail.com}
\orcid{0009-0008-1232-1190}
\affiliation{%
  \institution{School of computer science and engineering, Central South University}
  \city{Hunan}
  \country{China}
}

\author{Guangpeng Qi}
\email{qigp@inspur.com}
\orcid{0009-0006-4733-8975}
\affiliation{%
  \institution{Inspur Yunzhou Industrial Internet Co., Ltd}
  \city{Shandong}
  \country{China}
}
\author{Songtao Pang}
\email{pangst@inspur.com}
\orcid{0009-0004-4192-5568}
\affiliation{%
  \institution{Inspur Yunzhou Industrial Internet Co., Ltd}
  \city{Shandong}
  \country{China}
}
\author{Guangyong Shang}
\email{shangguangyong@inspur.com}
\orcid{0009-0000-8571-3605}
\affiliation{%
  \institution{Inspur Yunzhou Industrial Internet Co., Ltd}
  \city{Shandong}
  \country{China}
}

\author{Qiang Ma}
\email{tsinghuamq@gmail.com}
\orcid{0000-0001-5791-1890}
\authornote{Qiang Ma is the corresponding author.}
\affiliation{%
  \institution{Tsinghua University}
  \city{Beijing}
  \country{China}
}

\author{Zheng Yang}
\email{hmilyyz@gmail.com}
\orcid{0000-0003-4048-2684}
\affiliation{%
  \institution{School of Software, Tsinghua University}
  \city{Beijing}
  \country{China}
}


\renewcommand{\shortauthors}{Danyang Li et al.}


\begin{abstract}
Grounding natural language instructions to visual observations is fundamental for embodied agents operating in open-world environments. Recent advances in visual-language mapping have enabled generalizable semantic representations by leveraging vision-language models (VLMs). However, these methods often fall short in aligning free-form language commands with specific scene instances, due to limitations in both instance-level semantic consistency and instruction interpretation.
We present \textbf{OpenMap}, a zero-shot open-vocabulary visual-language map designed for accurate instruction grounding in navigation tasks. To address semantic inconsistencies across views, we introduce a \textit{Structural-Semantic Consensus} constraint that jointly considers global geometric structure and vision-language similarity to guide robust 3D instance-level aggregation. To improve instruction interpretation, we propose an LLM-assisted \textit{Instruction-to-Instance Grounding} module that enables fine-grained instance selection by incorporating spatial context and expressive target descriptions.
We evaluate \sysname on ScanNet200 and Matterport3D, covering both semantic mapping and instruction-to-target retrieval tasks. Experimental results show that \sysname outperforms state-of-the-art baselines in zero-shot settings, demonstrating the effectiveness of our method in bridging free-form language and 3D perception for embodied navigation.
\end{abstract}








\maketitle
 
\section{Introduction}
\label{sec:intro}

\begin{figure*}
  \centering
  \setlength\abovecaptionskip{4pt}
  \captionsetup[subfloat]{skip=0pt}
    \subfloat[Instance-level semantic mapping.]{
      \includegraphics[height=0.195\textwidth]{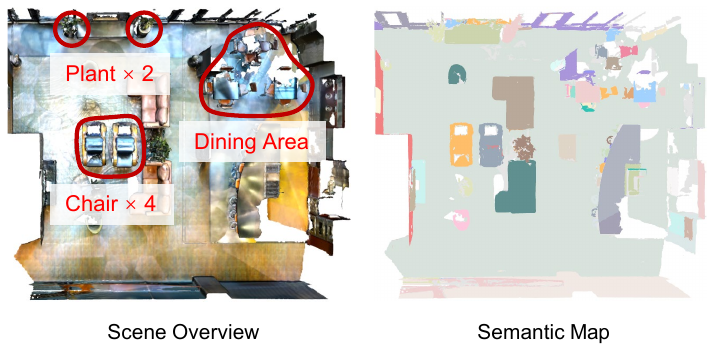}
      \label{fig:intro_1}
    }
    \subfloat[Grounding generic navigation instructions to target instances.]{
      \includegraphics[height=0.195\textwidth]{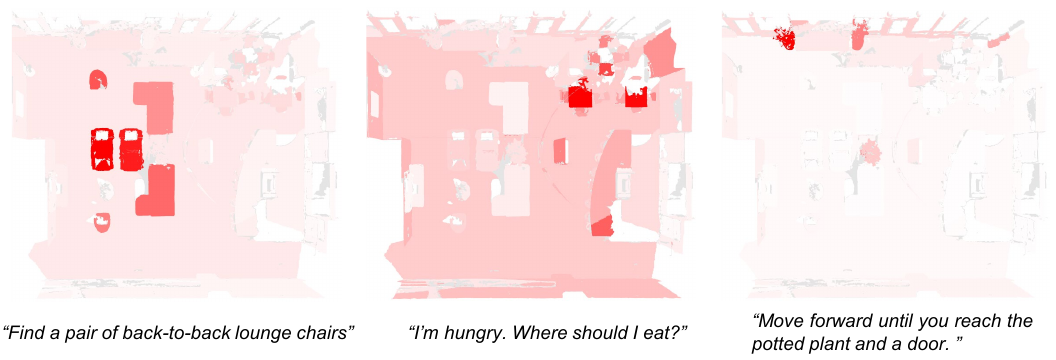}
      \label{fig:intro_2}
    }
    \caption{\sysname constructs an open-vocabulary visual-language map. \textnormal{(a) \sysname performs fine-grained, instance-level semantic mapping on navigation scenes from Matterport3D \cite{yadav2023habitat}. (b) Three types of navigation instructions are shown from left to right: object-goal, demand-driven, and language-guided. \sysname accurately grounds generic instructions to the intended targets, where darker regions in the heatmaps indicate stronger alignment between the instruction and the predicted instance.}}
  \label{fig:intro}
\end{figure*}

As the field of embodied intelligence continues to evolve, enabling agents to navigate using natural language instructions has emerged as a core challenge \cite{duan2022survey}.  
In Vision-and-Language Navigation (VLN), agents are expected to interpret language commands and perform goal-directed planning based on visual observations in complex 3D environments \cite{park2023visual, wu2024vision}.  
To support this process, semantic maps are essential, as they enhance perceptual understanding and enable precise, instruction-driven navigation \cite{takmaz2023openmask3d, wang2023gridmm}.  
Recent advances have further incorporated semantic features from VLMs \cite{radford2021learning, jia2021scaling} into 3D scene representations, giving rise to open-vocabulary visual-language maps \cite{huang2023visual, werby2024hierarchical, gu2024conceptgraphs} that generalize well across a wide range of navigation tasks.
However, effectively grounding natural language instructions to specific 3D instances within these maps—\ie, instruction grounding—remains an open challenge.

Existing open-vocabulary visual-language maps typically follow a two-stage pipeline:  
(1)~\textbf{Semantic mapping}: As agents navigate the environment, they collect visual observations and use VLMs to extract open-vocabulary features at the pixel level. These features are back-projected into 3D and aggregated across views to construct the semantic map.  
(2)~\textbf{Instruction grounding}: Given a free-form instruction, large language models (LLMs) generate descriptive target expressions, which are then grounded to the map by matching them against visual-language features.

While promising, these approaches still face significant limitations. In many cases, the alignment between natural language instructions and map instances underperforms even basic text-to-image matching capabilities of VLMs. Two core challenges remain:

\noindent$\bullet$ \textbf{In semantic mapping}, existing methods often rely on spatial or structural constraints to merge observations from different viewpoints. Some cluster point clouds by proximity or predefined grids \cite{chen2023open, huang2023visual}, which may lead to over- or under-segmentation. More advanced techniques leverage structural overlaps \cite{werby2024hierarchical, yan2024maskclustering}, but incomplete point clouds and object occlusion can still cause erroneous merges between semantically distinct instances.

\noindent$\bullet$ \textbf{In instruction grounding}, despite operating on open-vocabulary features, most methods remain tied to predefined object lexicons. Some rely on static category labels \cite{werby2024hierarchical, gu2024conceptgraphs}, while others restrict LLM outputs to a fixed set of terms \cite{huang2024ivlmap, huang2023visual}. 
These constraints hinder the expressive capacity of LLMs and limit their ability to capture fine-grained, contextualized object references. For example, given the instruction \textit{"Get the chair ready—I want to eat"}, identifying the chair near the dining table, rather than a generic chair, is non-trivial.

In summary, existing visual-language maps—across both semantic mapping and instruction grounding—largely follow closed-vocabulary paradigms, limiting the potential of VLMs and LLMs in open-world navigation. On one hand, they fail to fully exploit structural and semantic cues for robust instance association; on the other, they underutilize the generative flexibility of LLMs in grounding diverse instructions.

\textbf{Our Work.}  
We introduce \textbf{\sysname}, a zero-shot \textbf{Open} vocabulary visual-language \textbf{Map} designed for accurate instruction grounding in embodied navigation.  
\sysname addresses the above challenges by aligning natural language instructions with 3D instances through a unified visual-language representation. Specifically, we propose a structural-semantic consensus constraint that leverages both geometric and semantic consistency to drive robust instance merging during mapping (\fig\ref{fig:intro_1}).  
Furthermore, we introduce an instruction-to-instance grounding module that allows LLMs to generate fine-grained target descriptions and reason over spatial context for precise grounding (\fig\ref{fig:intro_2}).
\sysname offers key advantages in two core aspects:

\noindent$\bullet$ We propose a \textit{structural-semantic consensus mapping} strategy (\S\ref{sec:design1}) to resolve feature inconsistencies commonly introduced during the aggregation stage of existing mapping methods. Our approach incrementally constructs a 3D instance-level semantic map from 2D masks, using structural and semantic consensus as joint criteria for observation fusion. Specifically, two masks are merged only when supported by both global structural and semantic consistency—that is, they are mutually observable from other viewpoints (\ie, exhibit containment relationships in the point cloud) and closely aligned in the vision-language feature space (\ie, refer to the same object or its constituent parts). Guided by these consensus cues, \sysname iteratively aggregates 2D instances across views, effectively extending 2D vision-language alignment to 3D instance-level representations.

\noindent$\bullet$ We introduce a \sysname-enhanced \textit{Instruction-to-Instance grounding} module (\S\ref{sec:design2}). Unlike prior approaches that constrain LLM outputs using a predefined scene instance lexicon when interpreting navigation instructions \cite{huang2023visual, long2024instructnav, werby2024hierarchical}, \sysname enables more precise grounding of natural language to scene instances. This allows large language models to generate fine-grained instance descriptions—for example, interpreting “\textit{I am thirsty}” as “\textit{a cup filled with water}” rather than simply “\textit{cup}.” Furthermore, \sysname incorporates spatial context to support reasoning over candidate instances; for instance, given the instruction “\textit{Get the chair ready—I want to eat},” it can identify the intended chair by considering nearby objects such as a dining table. The synergy between LLMs and \sysname enables accurate indexing from high-level navigation instructions to specific map instances.

We evaluate \sysname on the public benchmark ScanNet200 \cite{rozenberszki2022language}, focusing on instance segmentation precision and semantic accuracy. To further assess its target retrieval capabilities in navigation scenarios, we conduct experiments on the Matterport3D dataset \cite{yadav2023habitat} using a variety of instruction types, including object-goal \cite{sun2024survey}, demand-driven \cite{wang2023find}, and language-guided instructions \cite{krantz2020beyond}. Compared to state-of-the-art (SOTA) methods, \sysname consistently outperforms them in both zero-shot semantic mapping and instruction-to-target grounding.

\begin{figure*}[t]
  \centering
  \setlength\abovecaptionskip{1pt}
  \includegraphics[width=0.96\textwidth]{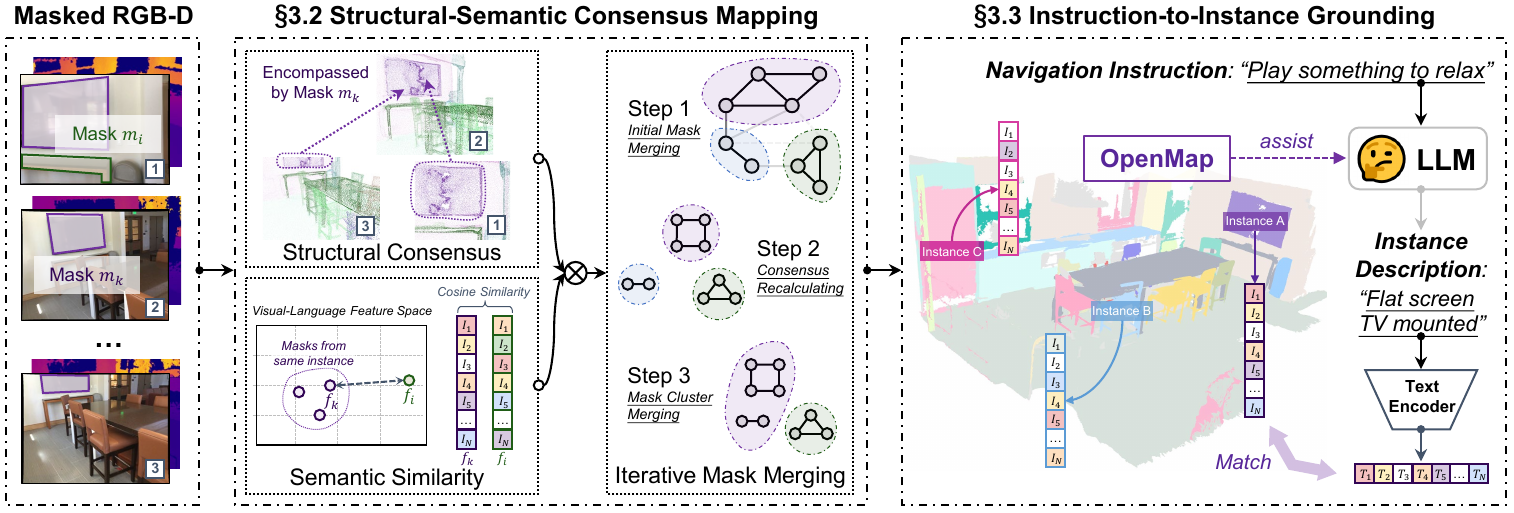}
  \caption{\sysname Overview. \textnormal{\sysname takes RGB-D inputs from multiple viewpoints and applies pretrained models to predict 2D masks and extract open-vocabulary features. During semantic mapping (\S\ref{sec:design1}), it iteratively aggregates 2D masks into 3D instances using a structural-semantic consensus constraint. During instruction grounding  (\S\ref{sec:design2}), an LLM selects the target instance by reasoning over candidate proposals and scene context provided by \sysname.}}
  \label{fig:overview}
\end{figure*}

Our contributions are summarized as follows:

\noindent $\bullet$ We develop \textbf{\sysname}, an open-vocabulary visual-language mapping framework that bridges LLM-based instruction parsing and 3D instance grounding, enabling precise instruction grounding from free-form navigation commands.

\noindent $\bullet$ We propose a novel structural-semantic consensus constraint that jointly leverages global geometric consistency and vision-language semantics to enable fine-grained 3D instance-level mapping.

\noindent $\bullet$ We evaluate \sysname on ScanNet200 and Matterport3D, covering both semantic mapping and target retrieval tasks, and show consistent improvements over existing methods. \textit{Our code is publicly available at https://github.com/openmap-project/OpenMap}.

\section{Related Work}
\label{sec:related_work}

\noindent\textbf{Vision-Language Foundation Models.}  
Large-scale VLMs, such as CLIP~\cite{radford2021learning} and BLIP~\cite{li2022blip}, align visual and textual modalities within a shared embedding space~\cite{zhang2024vision}. These advances have facilitated open-vocabulary understanding~\cite{wu2024towards} across tasks such as classification and retrieval.
Recent efforts extend these capabilities to 2D segmentation, with models like OVSeg~\cite{liang2023open} and OVSAM~\cite{yuan2024open} incorporating segmentation heads to support instance-level open-vocabulary queries~\cite{zhu2024survey}. However, transferring this alignment into 3D space remains challenging due to sparse and incomplete data, especially in navigation settings where environments are incrementally explored~\cite{huang2023clip2point, chen2023clip2scene}. 

\noindent\textbf{Open-Vocabulary 3D Instance Mapping.} 
Among various forms of open-vocabulary 3D semantic mapping, instance-level mapping is particularly challenging yet essential for accurate instruction grounding.
Recent methods for open-vocabulary 3D instance segmentation~\cite{lai2023mask,nguyen2024open3dis} follow two main paradigms. The first, \textit{3D-to-2D} \cite{takmaz2023openmask3d, huang2024openins3d, peng2023openscene}, performs segmentation in 3D and projects results to 2D for feature extraction, but often suffers from poor completeness and semantic consistency due to sparse point clouds. The second, \textit{2D-to-3D}~\cite{gu2024conceptgraphs, lu2023ovir, yan2024maskclustering}, segments 2D frames, uses depth maps for 3D back-projection, and aggregates instance masks via geometric overlap or spatial clustering. While effective, these methods typically overlook semantic similarity in the merging process. 

\noindent\textbf{Semantic Mapping for Instruction Grounding.} 
Semantic maps are critical in VLN, as they allow agents to reason over spatial and semantic structures for instruction interpretation and execution~\cite{wu2024vision, huang2023visual}. Early works project 2D instance masks onto bird’s-eye-view layouts~\cite{fan2024navigation, liu2023bird}, or aggregate features into top-down grids to support open-vocabulary querying~\cite{wang2023gridmm, huang2024ivlmap}. VLMap~\cite{huang2023visual} introduces semantic grid maps and uses LLMs to translate instructions into open-vocabulary object names. 
ConceptGraphs~\cite{gu2024conceptgraphs} and HOV-SG~\cite{werby2024hierarchical} further enhance instruction interpretation by constructing spatial graphs over scene instances, enabling explicit modeling of inter-object relationships. 
However, these methods still rely on predefined labels or coarse semantic features, limiting their ability to support fine-grained, open-vocabulary instruction grounding. 
\section{Methodology}
\label{sec:methodology}

\subsection{Method Overview}
\label{sec:overview}

An overview of \sysname is shown in \fig\ref{fig:overview}. 
We follow a generic agent setting for embodied localization \cite{xu2020edge,li2024edgeslam2} and navigation~\cite{huang2023visual, long2024instructnav}, where an agent collects a sequence of RGB-D observations during exploration \cite{li2022motion,li2023leovr}, denoted as $\mathcal{I} = \{ I_1, I_2, ..., I_T \}$ and $\mathcal{D} = \{ D_1, D_2, ..., D_T \}$. For each frame $I_t$, we apply an off-the-shelf 2D segmentation model to generate masks $\{m^t_i \mid i = 1, ..., n_t\}$ and use a vision-language model to extract corresponding open-vocabulary features $\{f^t_i \mid i = 1, ..., n_t\}$, where $n_t$ is the number of masks in $I_t$.

During the semantic mapping stage (\S\ref{sec:design1}), we apply structural and semantic consensus constraints to determine whether any two masks across the image sequence correspond to the same instance, and iteratively merge those that satisfy both into a unified 3D instance. We then adopt a completeness-guided strategy to select representative masks and aggregate their features to form a holistic semantic embedding for each instance.

In the instruction grounding stage (\S\ref{sec:design2}), the agent receives a natural language instruction and leverages an LLM to parse it into a target instance description. Unlike conventional methods constrained by predefined vocabularies, our approach allows for free-form, fine-grained descriptions of navigation goals. Guided by \sysname, the LLM selects the most semantically relevant instance by reasoning over candidate regions and their associated features within the constructed map.
\subsection{Structural-Semantic Consensus Mapping}
\label{sec:design1}

The core philosophy behind constructing \sysname is to concurrently consider both spatial structure and semantic feature constraints across different observations.

To visualize this concept, consider a scenario where three observers (\eg, $O_A$, $O_B$, and $O_C$) are examining the same object (\eg, a bunch of roses) from distinct angles. 
How should they describe it to ascertain that they are indeed looking at the same object?

\noindent $\bullet$ Semantically, the observations should exhibit similarities—for instance, $O_A$ might note `\textit{a few green leaves}', $O_B$ could describe `\textit{three roses}', and $O_C$ might see a `\textit{bunch of flowers}'. 

\noindent $\bullet$ Structurally, there should be a consensus, such that the parts of the instance observed by $O_A$ and $O_B$ are also encompassed in $O_C$’s observation, suggesting that these observations originate from the same instance. 

We proceed by modeling these structural-semantic consensus constraints to facilitate precise instance merging.

\subsubsection{\textbf{Structural-Semantic Consensus Rate Computing.}}
\label{sec:design1_rate}

For any two masks \(m_i\) and \(m_j\), we evaluate their potential for merging by assessing the structural-semantic consensus rate between them. 

\noindent \textbf{Structural Consensus Rate.}
Drawing on well-established structural consensus analysis \cite{triggs2000bundle,xu2022swarmmap}, we strategically harness the redundancy of observations to ensure structural self-consistency of instances
Specifically, following \cite{yan2024maskclustering}, we back-project each mask \(m_k\) into a 3D point cloud \(P_k\) using depth maps. If the point cloud of an image \(I\) overlaps with \(P_k\) (\ie,  overlap exceeds $\tau_{obs}$), then mask \(m_k\) is deemed observable in image \(I\), and we define \(\mathcal{I}(m_k)\) as the set of all images that can observe mask \(m_k\).

We then identify the set of images that can simultaneously observe both masks \(m_i\) and \(m_j\) intended for merging, denoted as \(O(m_i, m_j) = \mathcal{I}(m_i) \cap \mathcal{I}(m_j)\).
Subsequently, we seek frames capable of supporting the merger of \(m_i\) and \(m_j\). Specifically, for an image \(I_t\) containing a mask \(m_k\) that spatially encompasses both \(m_i\) and \(m_j\) (\ie, both have at least \(\tau_{sub}\) of their point clouds within \(m_k\)), this subset of images is defined as \(S(m_i, m_j) = \{I_t \in O(m_i, m_j) | P_i, P_j \subseteq P_k\}\).
Consequently, the structural consensus rate for the two masks \(m_i\) and \(m_j\) is calculated as the ratio of supporters to observers:
\begin{equation}
    R_{struc.}(m_i, m_j) = {|S(m_i, m_j)|}/{|O(m_i, m_j)|}.
    \label{eq:rate_structural}
\end{equation}

\begin{figure}[t]
  \centering
  \setlength\abovecaptionskip{2pt}
  \includegraphics[width=0.46\textwidth]{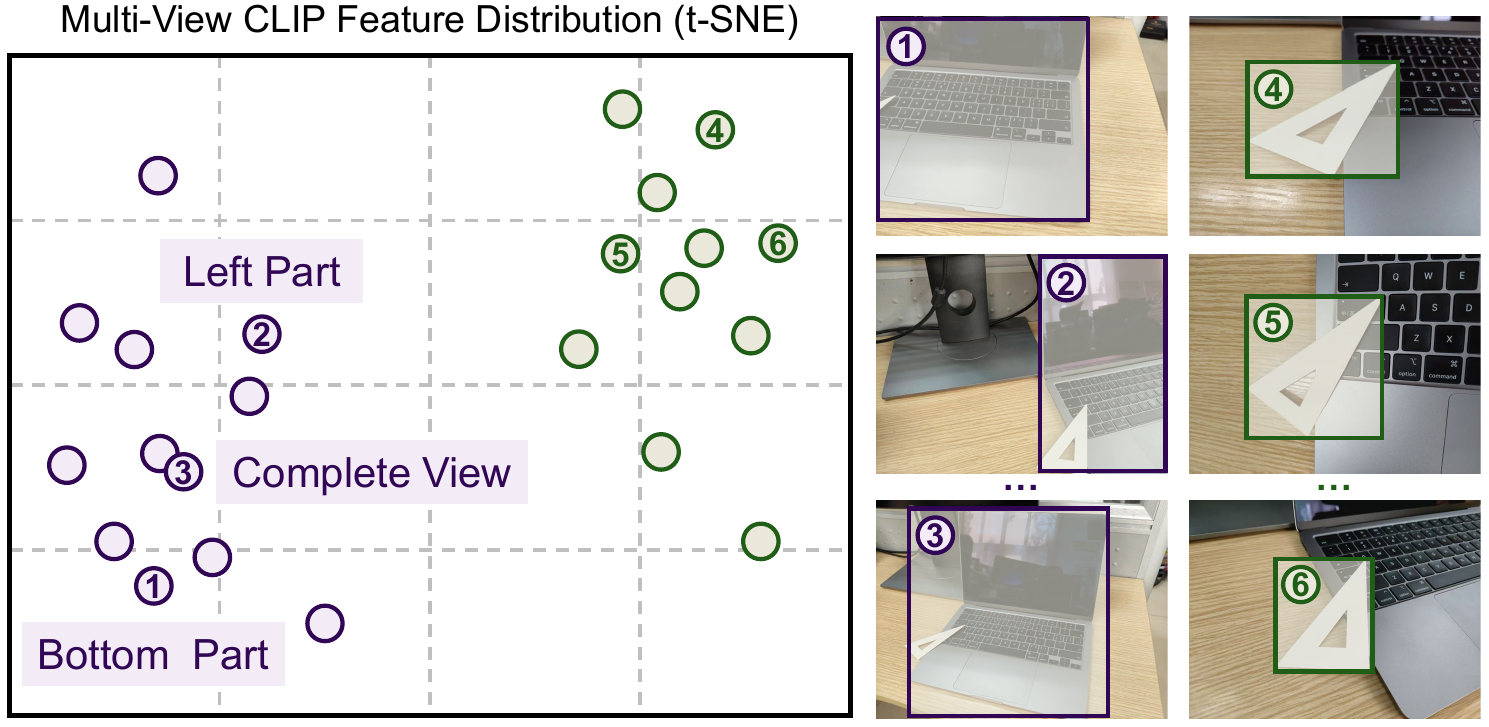}
  \caption{Feature distribution of adjacent objects. \textnormal{Although the triangle ruler is physically attached to the laptop, it remains clearly distinguishable in the vision-language feature space.}}
  \label{fig:tSNE}
\end{figure}

\noindent \textbf{Semantic Similarity Rate.}
Another criterion for determining whether masks can be merged is their semantic similarity. 
Diverging from traditional models that rely on a closed vocabulary, VLMs like CLIP capture subtle semantic connections between observations, even different components of the same instance.
As shown in \fig\ref{fig:tSNE}, features from the same instance are tightly clustered in the latent space, while spatially close but distinct instances exhibit clear feature separation.

Consequently, by integrating open-vocabulary feature similarity metrics into instance merging, we can effectively reduce the misalignment of different instances that are close in spatial structure but semantically distinct.
We define the semantic similarity rate between masks \(m_i\) and \(m_j\) as the cosine similarity between their respective features \(f_i\) and \(f_j\):

\begin{equation}
    R_{seman.}(m_i, m_j) = \cos(f_i, f_j).
    \label{eq:rate_semantic}
\end{equation}

\noindent \textbf{Put Together.}
In mask merging, we balance the above structural consensus and semantic similarity. Specifically, when 
\begin{equation}
    R_{struc.}(m_i, m_j) * R_{seman.}(m_i, m_j) \geq \tau_{thres},
    \label{eq:rate_all}
\end{equation}
masks \(m_i\) and \(m_j\) are considered to form the same instance, where \(\tau_{thres}\) is a predefined threshold.

\subsubsection{\textbf{Iterative Mask Merging}}
\label{sec:design1_merge}

After computing pairwise relationships between masks, we iteratively merge them following the general procedure in \cite{yan2024maskclustering}. This results in a set of 3D point clouds, each representing a distinct instance, with open-vocabulary semantic features aggregated from the corresponding masks.

Specifically, we prioritize merging mask pairs associated with more robust observations (\ie, a larger \(|O(m_i, m_j)|\)). Therefore, during the iterative process, we set a gradually decreasing threshold for observer counts, \(N_o\). In each merging cycle, two masks, \(m_i\) and \(m_j\), are merged into a new mask, \(m_{i,j}\), with its corresponding point cloud, \(P_{i,j}\), if they not only meet the conditions set by \eq\ref{eq:rate_all} but also exceed the observer count threshold, \(|O(m_i, m_j)| > N_o\).

After each merging cycle, it's necessary to recalculate the structural semantic consensus rate among the newly formed instances. The strategy is as follows: 

\noindent $\bullet$ Given the structural changes in the new instance, along with altered observational and containment relationships, we update its observers and supporters and recompute \(R_{struc.}\).

\noindent $\bullet$ Furthermore, due to the aggregation of diverse observations, the semantic features of the combined instance need to be updated. We select the features from the mask that most completely captures the point cloud of the new instance and recalculate \(R_{seman.}\).

Each iteration cycle reduces threshold \(N_o\), allowing the instance to incorporate more masks, and this process continues until no further merging is feasible. At the end of the iterations, a list of 3D instances is generated, each linked to multiple 2D masks.

Finally, based on the completeness of the instance's observation, we strategically aggregate features for each instance. Following OpenMask3D \cite{takmaz2023openmask3d}, we select the top-$k$ masks that best cover the instance and obtain $L$ multi-level crops from the corresponding image areas. Subsequently, features are extracted from these $k*L$ crops using CLIP, with the average pooling results serving as the open-vocabulary feature vector for the instance.


\subsection{Instruction-to-Instance Grounding}
\label{sec:design2}

In everyday life, when colleagues hand over a task, from the arranger's perspective, it is essential to describe the requirements as accurately as possible, rather than being vague. From the executor's perspective, any unclear requirements should be clarified based on contextual information to eliminate ambiguities.

These experiences are equally applicable to embodied navigation tasks, and the parsing of instructions to instances adheres to the following principles:

\noindent $\bullet$ When utilizing LLMs for instruction translation, the instructions should be converted into instances described precisely in natural language, instead of choosing from a limited set of dataset labels or an instance dictionary specific to a certain scenario.

\noindent $\bullet$ Even with accurate instance descriptions, there may be multiple suitable targets in the scene, and agents should enhance their decision-making by considering contextual information such as the locations of candidates and their surrounding environment.

Next, we will discuss how to utilize \sysname to assist LLMs in implementing these concepts.

\noindent \textbf{Generic Instruction Parsing.}
We first utilize an LLM to convert generic navigation instructions into precise instance descriptions that enable target retrieval within \sysname. Unlike existing methods that translate various types of navigation instructions into targets from a fixed instance dictionary \cite{long2024instructnav}, our approach fundamentally differs in that we do not restrict the LLM outputs to predefined dictionary terms, thus fully unleashing its vast knowledge and analytical capabilities. 

The textual prompt is composed of the following key components:
(1) \textit{Navigation Task Definition}: Similar to existing work in navigation instruction analysis, we provide the background of the navigation task, including environmental information and the format of the instructions.
(2) \textit{Instance Description Criteria}: We retain the descriptions of instance characteristics from the original navigation instructions and infer additional features based on environmental information to reduce linguistic ambiguities.
(3) \textit{Output Format Constraints}: We streamline the instance descriptions to avoid excessive length, given that most existing VLMs (e.g., the original CLIP) have limited capabilities to process long and complex texts.

For an instance description derived from a navigation instruction, where the VLM encodes the feature vector as \(f_l\), we calculate the similarity with all instance features in \sysname \(\{f_1, f_2, ..., f_N\}\) and rank them. The top \(N_c\) are selected as candidate instances.

\begin{figure}[t]
  \centering
  \setlength\abovecaptionskip{1pt}
  \includegraphics[width=0.46\textwidth]{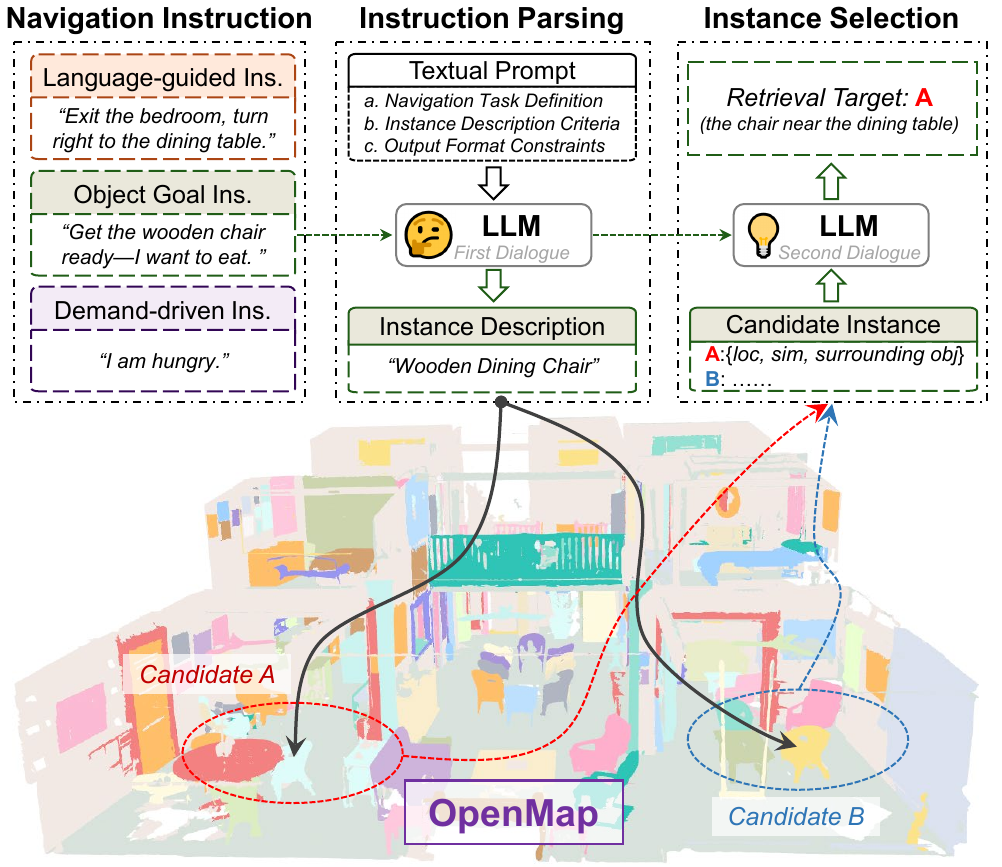}
  \caption{Instruction-to-Instance Grounding pipeline. \textnormal{In the first round, the LLM generates a precise description of the navigation target and retrieves candidate instances from \sysname. In the second round, it reasons over the candidates and their surrounding context to infer the final target instance.}}
  \label{fig:LLM}
\end{figure}

\noindent \textbf{Instance Selection with \sysname.}
Due to the potential presence of multiple instances in a scene that closely match the target description, relying solely on semantic similarity often fails to provide accurate measurements.
However, leveraging environmental constraints around candidate instances can significantly enhance the performance of targeted retrieval. A typical case is "\textit{Prepare the chair, I want to eat}". Typically, chairs can found in every room, but the follow-up prompt "\textit{I want to eat}", implies that the chair needed is one near the dining table.

For \(N_c\) candidates, we further refine our selection using the instance information provided by \sysname. Specifically, for each candidate instance \(\texttt{I}_k\), we use a KD-tree to search within \sysname for the \(N_s\) nearest instances \(\texttt{I}^i_k\) (i = 1, 2, ..., \(N_s\)) within a 2-meter radius. Subsequently, we label these \(\texttt{I}^i_k\) (e.g., match with the labels from the LVIS dataset \cite{gupta2019lvis}). Note that applying fixed labels here is solely to assist the LLM in instance selection and does not compromise the open-vocabulary querying capabilities of \sysname.

Next, we initiate a second round of dialogue with the LLM, providing details about the candidate instance and its surrounding objects, and determining the final retrieval target through a multiple-choice format.
The template for providing instance information in the prompt is abstracted as follows:

\noindent 
"Candidate Instance \(\texttt{I}_k\): \{location: (\(x_k\), \(y_k\), \(z_k\)); semantic similarity: \(r_k\); surrounding objects: [\(\texttt{I}^i_k\), location: (\(x^i_k\), \(y^i_k\), \(z^i_k\)), label: \(l^i_k\)], ...\}"

\begin{table*}[t]
\centering
\renewcommand{\arraystretch}{1}
\begin{tabular}{lccccccccccc}
\toprule
\multirow{2}{*}{\textbf{Model}} & \multirow{2}{*}{\textbf{Features}} & \multicolumn{6}{c}{\textbf{Semantic}} & \multicolumn{3}{c}{\textbf{Class-agnostic}} \\
\cmidrule(lr){3-8} \cmidrule(lr){9-11}
 & & AP & AP$_{50}$ & AP$_{25}$ & head(AP) & common(AP) & tail(AP) & AP & AP$_{50}$ & AP$_{25}$ \\
\midrule
\textit{sup. mask + sup. semantic} & & & & & & & & & & \\
 Mask3D \cite{schult2023mask3d} & -- & 26.9 & 36.2 & 41.4 & 39.8 & 21.7 & 17.9 & 39.7 & 53.6 & 62.5 \\
\cmidrule(lr){1-11}
\textit{sup. mask + z.s. semantic} & & & & & & & & & & \\
 OpenScene \cite{peng2023openscene} + Masks & OpenSeg \cite{ghiasi2022scaling} & 11.7 & 15.2 & 17.8 & 13.4 & 11.6 & 9.9 & 39.7 & 53.6 & 62.5 \\
 OpenMask3D \cite{takmaz2023openmask3d} & CLIP \cite{radford2021learning} & 15.4 & 19.9 & 23.1 & 17.1 & 14.1 & 14.9 & 39.7 & 53.6 & 62.5 \\
\cmidrule(lr){1-11}
\textit{z.s. mask + z.s. semantic} & & & & & & & & & & \\
 OVIR-3D \cite{lu2023ovir} & CLIP \cite{radford2021learning} & 9.3 & 18.7 & 25.0 & 10.1 & 9.4 & 8.1 & 14.4 & 27.5 & 38.8 \\
 MaskClustering \cite{yan2024maskclustering} & CLIP \cite{radford2021learning} & 12.0 & 23.3 & 30.1 & 11.9 & 10.5 & 13.8 & 19.0 & 36.6 & 50.8 \\
 \textbf{\sysname (Ours)} & CLIP \cite{radford2021learning} & \textbf{14.3} & \textbf{26.0} & \textbf{33.3} & \textbf{14.5} & \textbf{13.8} & \textbf{14.7} & \textbf{19.8} & \textbf{38.0} & \textbf{51.8} \\
\bottomrule
\end{tabular}
\caption{3D Instance Segmentation Results on ScanNet200 \cite{rozenberszki2022language}. \textnormal{Mask3D \cite{schult2023mask3d} requires supervised (\textit{sup.}) training on ScanNet200 for mask and semantic extraction. OpenScene \cite{peng2023openscene} + Masks and OpenMask3D \cite{takmaz2023openmask3d} depend on masks provided by Mask3D. In a fully zero-shot (\textit{z.s.}) setting, our method, \sysname, surpasses both OVIR-3D \cite{lu2023ovir} and MaskClustering \cite{yan2024maskclustering} across all metrics.}}
\label{table:map}
\end{table*}

\begin{table}[t]
\centering
\begin{tabular*}{\linewidth}{@{\extracolsep{\fill}}lcccc}
\toprule
\textbf{Method} & SR[\%] & SR$_{4}$[\%] & SR$_{8}$[\%] & SR$_{16}$[\%] \\
\midrule
NLMap \cite{chen2023open} & 25.1 & 28.4 & 31.5 & 37.1 \\
VLMap \cite{huang2023visual}  & 27.2 & 29.7 & 32.1 & 37.5 \\
ConceptGraphs \cite{gu2024conceptgraphs} & 40.9 & 43.4 & 50.6 & 54.9 \\
\textbf{\sysname (Ours)} & \textbf{49.6} & \textbf{58.3} & \textbf{68.2} & \textbf{73.7} \\
\bottomrule
\end{tabular*}
\caption{Navigation target retrieval results on Matterport3D \cite{yadav2023habitat}. \textnormal{Compared with existing open-vocabulary mapping methods designed for navigation tasks, \sysname\ achieves the best performance across all target retrieval success rate metrics.}}
\label{table:nav}
\end{table}

\section{Experiments}
\label{sec:experiments}
In this section, we evaluate \sysname against current SOTA methods in terms of semantic mapping and target retrieval. 

\subsection{Experimental Setup}
\label{sec:setup}

\noindent \textbf{Dataset.}
We utilize the ScanNet200 validation dataset \cite{rozenberszki2022language} to evaluate \sysnameposs semantic mapping capabilities. This dataset features 312 indoor scans across 200 categories, organized into three subsets based on the frequency of instance occurrences, allowing for an effective assessment across a long-tail distribution.
Additionally, we examine \sysnameposs navigation target retrieval effectiveness with the Matterport3D Semantics dataset \cite{yadav2023habitat}, following established navigation map research \cite{huang2023visual}\cite{werby2024hierarchical}. 
Our evaluation spans 20 scenes—11 from the R2R-CE val-unseen split~\cite{krantz2020beyond} and 9 from the VLMap evaluation dataset~\cite{huang2023visual}. We construct a comprehensive set of test cases using subsets of navigation instructions from R2R-CE, VLMap, and ALFRED~\cite{shridhar2020alfred}, covering three instruction types: object-goal, demand-driven, and language-guided. 

\vspace{2pt}

\noindent \textbf{Baselines.}
We evaluated \sysname against SOTA 3D semantic mapping and VLN target retrieval methods.
For semantic mapping, \textbf{Mask3D} \cite{schult2023mask3d} is a representative work trained under supervision on ScanNet200.
\textbf{OpenScene} \cite{peng2023openscene} is an open-vocabulary 3D scene understanding model that generates per-point feature vectors, for which we average the per-point features within each instance mask, following the approach in \cite{takmaz2023openmask3d}.
\textbf{OpenMask3D} \cite{takmaz2023openmask3d} utilizes supervised mask proposals from Mask3D and employs CLIP for open-vocabulary semantic aggregation.
\textbf{OVIR-3D} \cite{lu2023ovir} and \textbf{MaskCLustering} \cite{yan2024maskclustering} are zero-shot open-vocabulary mapping methods that aggregate instances progressively from 2D to 3D, closely related to our approach.
For target retrieval, \textbf{NLMap} \cite{chen2023open} and \textbf{VLMap} \cite{huang2023visual} utilize LLMs to retrieve targets of known categories on constructed queryable maps. 
\textbf{ConceptGraphs} \cite{gu2024conceptgraphs} and \textbf{HOV-SG} \cite{werby2024hierarchical} further enhance the object retrieval and reasoning capabilities by constructing graphs between instances.

\vspace{2pt}

\noindent \textbf{Metrics.}
To validate our mapping accuracy, we report Average Precision (AP) at 25\% and 50\% Intersection over Union (IoU) thresholds, along with the mean AP from 50\% to 95\% at 5\% intervals. We also evaluate performance in a class-agnostic setting that focuses solely on mask quality, ignoring semantic labels.
For target retrieval in \sysname, we use the Success Rate (SR), defined as successful if the target is retrieved within 1 meter of the ground truth center. Additionally, we measure the top-$k$ Success Rate (SR$_k$), indicating success within up to $k$ retrieval attempts in the scene.

\vspace{2pt}

\noindent \textbf{Implementation Details.}
To obtain complete object masks rather than overly fragmented results (\ie, all pixels of an object belonging to one mask), we employ CropFormer \cite{qilu2023high,qi2023high} for 2D segmentation. An intuitive approach for encoding visual-language features for each mask involves cropping bounding boxes and extracting features using CLIP. However, this process generates an excessive number of image patches and, limited by CLIP's processing speed, proves inefficient. We utilize OVSAM \cite{yuan2024ovsam} to extract features for candidate regions within an image in one go, which is achieved by using the bounding boxes of these regions as prompts for OVSAM. Note that OVSAM features are only used for computing the semantic similarity rate. For a fair comparison of semantic matching capabilities with existing methods, we use features extracted by CLIP \cite{radford2021learning} ViT-H for the final feature aggregation. We apply post-processing methods from MaskCLustering to filter under-segmented masks and separate disconnected point clusters into distinct instances.
Regarding parameter settings, in \S\ref{sec:design1_rate}, the observational threshold for masks \(\tau_{obs} = 0.3\), the containment threshold for masks \(\tau_{sub} = 0.8\), and the  threshold of structural-semantic consensus rate \(\tau_{thres} = 0.6\); in \S\ref{sec:design1_merge}, the initial threshold for the number of observers \(N_o\) is set at the top 5\% of all mask pairs, reducing by 5\% in each iteration until the process concludes; in \S\ref{sec:design2}, the number of candidate instances \(N_c = 8\) and the number of neighboring instances \(N_s = 5\).

\begin{figure*}[t]
  \centering
  \setlength\abovecaptionskip{2pt}
  \includegraphics[width=0.96\textwidth]{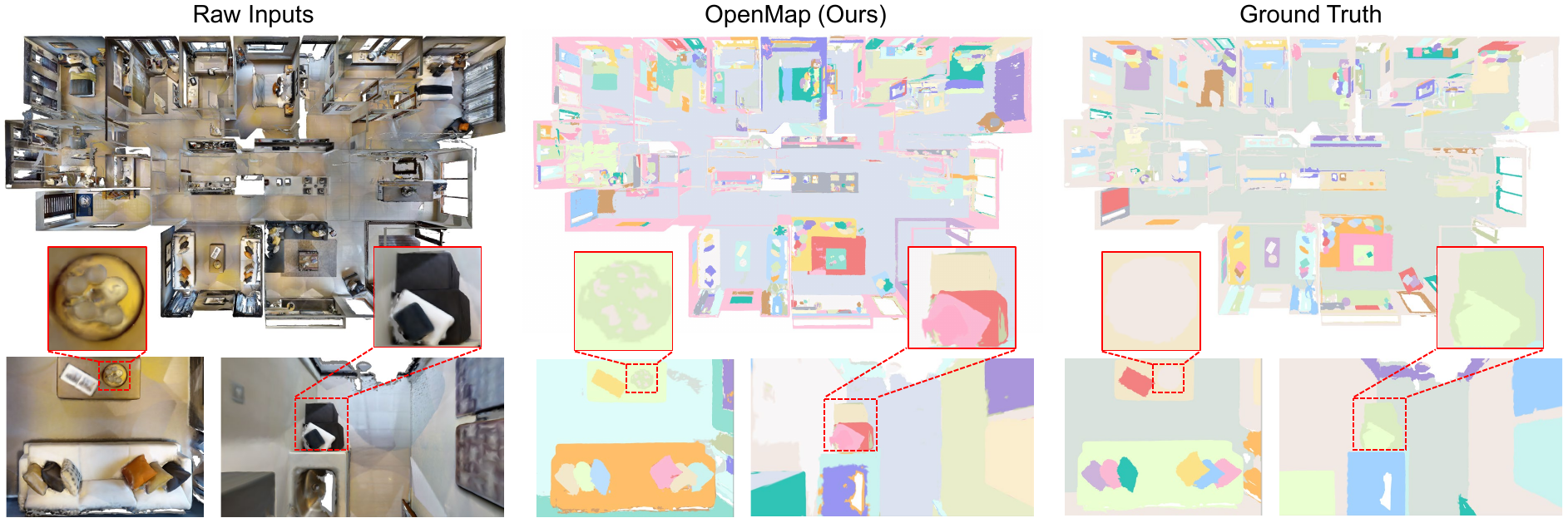}
  \caption{Semantic mapping results of \sysname on Matterport3D.}
  \label{fig:exp_show_1}
\end{figure*}

\begin{figure}[t]
  \centering
  \setlength\abovecaptionskip{2pt}
  \includegraphics[width=0.46\textwidth]{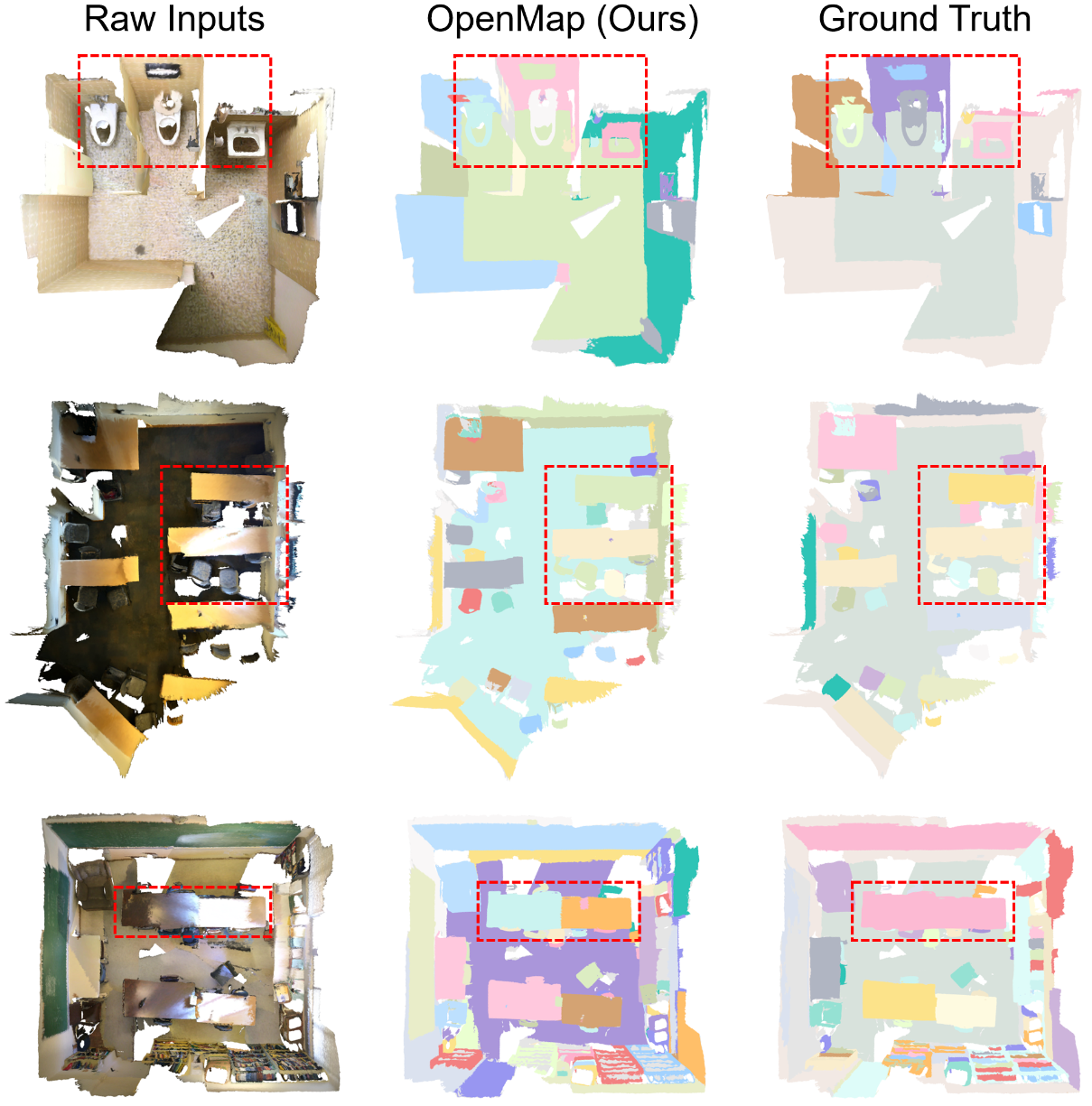}
  \caption{Semantic mapping results on ScanNet200.}
  \label{fig:exp_show2}
\end{figure}

\subsection{Mapping Performance}
\label{sec:exp_maping}

\noindent \textbf{Quantitative Results.}
Following standard practice in both supervised and zero-shot semantic mapping, we primarily report results on ScanNet200 as it serves as the most widely adopted benchmark, enabling fair comparison with prior work. As shown in \tab\ref{table:map}, we categorize the comparison methods into three groups.

Compared to the fully zero-shot OVIR-3D and MaskClustering, \sysname achieves the highest accuracy on ScanNet200 in both semantic and class-agnostic metrics. Specifically, \sysname shows a 19.2\% improvement in average semantic AP over MaskClustering, which only considers structural features during aggregation.
Furthermore, compared to OVIR-3D, which processes local geometric and semantic features frame by frame, \sysname, which incorporates global feature semantic consensus, shows even more pronounced improvements, with increases of 53.8\% in semantic AP and 37.5\% in class-agnostic AP.
Notably, compared to these zero-shot methods, \sysname maintains consistently high performance across head, common, and tail categories, with an AP gap not exceeding 1.1\%. This capability to handle instances of varying frequencies highlights \sysname's robust open-vocabulary abilities.

In contrast with OpenScene and OpenMask3D, \sysname, lacking any prior knowledge from ScanNet200, still shows a significant gap in the class-agnostic metrics. Nevertheless, \sysname significantly surpasses OpenScene in all semantic metrics due to its lack of strategic open-vocabulary feature aggregation. Moreover, \sysname is close to OpenMask3D in semantic AP, even exceeding it in AP50 and AP25 by +6.1\% and +10.2\%, respectively.
Additionally, OpenMask3D is a semantic mapping method based on 3D-to-2D projection, using a complete 3D point cloud of the scene as a prior. In navigation tasks where the scene is progressively explored, our method, \sysname, fits more seamlessly, able to integrate into existing navigation tools as a fundamental semantic map to support downstream tasks.

\noindent \textbf{Qualitative Results.}
As shown in \fig\ref{fig:exp_show_1} and \fig\ref{fig:exp_show2}, we present qualitative instance segmentation results of \sysname on Matterport3D and ScanNet200 scenes. \sysname demonstrates strong performance in two key scenarios: (1) accurately segmenting small objects attached to larger surfaces (\eg, scattered items on a table or glasses on a tray); and (2) preserving the completeness of large objects despite limited viewpoint coverage (\eg, preventing large sofas, tables, and beds from being mistakenly fragmented).  
Notably, in the first subview of the Matterport3D scene, \sysname successfully segments individual items on the tray, such as bowls and glasses. However, the ground truth merges these into a single instance, leading to a false negative during evaluation despite the correctness of the prediction.

\subsection{Target Retrieval Performance}
\label{sec:exp_navigation}
\noindent \textbf{Quantitative Results.}  
\sysname is designed to achieve accurate grounding of navigation instructions to scene instances, a task that jointly evaluates the quality of semantic mapping and the effectiveness of instruction-to-instance grounding.
We compare \sysname against two representative visual-language mapping baselines, NLMap \cite{chen2023open} and VLMap \cite{huang2023visual}, as well as the recent SOTA method ConceptGraphs \cite{gu2024conceptgraphs}, in terms of target retrieval success rate. For each trial, all methods first perform full-scene mapping, followed by instruction parsing via a LLM, and then query the target location based on the generated map. To ensure a fair comparison, all methods employ GPT-4 \cite{achiam2023gpt} as the LLM.

As shown in \tab\ref{table:nav}, \sysname significantly outperforms the baselines across all success rate metrics. In particular, under the SR metric—which reflects the most practical requirement in navigation (i.e., succeeding on the first attempt)—\sysname surpasses NLMap, VLMap, and ConceptGraphs by +24.5\%, +22.4\%, and +8.7\%, respectively.  
Among the baselines, VLMap suffers from a coarse feature aggregation strategy (i.e., average pooling over 2D grids), which fundamentally limits the quality of the underlying semantic map and thus its retrieval capability. ConceptGraphs, on the other hand, relies on pre-defined labels for instances, which restricts its generalization to diverse natural language descriptions.


\begin{table}[t]
\centering
\begin{tabular}{cc|ccc}
\toprule
\textbf{Structural.} & \textbf{Semantic.} & AP & AP$_{50}$ & AP$_{25}$ \\
\midrule
\ding{51} & \ding{55} & 12.2 & 23.4 & 30.2 \\
\ding{55} & \ding{51} & 10.1 & 19.3 & 26.7 \\
\ding{51} & \ding{51} & \textbf{14.3} & \textbf{26.0} & \textbf{33.3} \\
\bottomrule
\end{tabular}
\caption{Ablation study on Mapping methods.}
\label{table:ablation_map}
\end{table}

\begin{table}[t]
\centering
\begin{tabular}{cc|ccc}
\toprule
\textbf{Ins. Parsing} & \textbf{Ins. Selec.} & SR[\%] & SR$_{8}$[\%] & SR$_{16}$[\%] \\
\midrule
\ding{55} & \ding{55} & 38.1 & 61.0 & 67.6 \\
\ding{51} & \ding{55} & 47.2 & 68.2 & 73.7 \\
\ding{55} & \ding{51} & 44.7 & 61.0 & 67.6 \\
\ding{51} & \ding{51} & \textbf{49.6} & \textbf{68.2} & \textbf{73.7} \\
\bottomrule
\end{tabular}
\caption{Ablation study on Grounding methods.}
\label{table:ablation_nav}
\end{table}

\noindent \textbf{Qualitative Results.}  
\fig\ref{fig:intro} shows instance-colored segmentation results and similarity heatmaps generated by \sysname on a Matterport3D scene (ID: 8194nk5LbLH). As shown in \fig\ref{fig:intro_1}, the semantic map yields accurate instance-level segmentation for both common objects (\eg, sofas) and uncommon structures (\eg, columns). White regions in the overview indicate missing scan data.
In \fig\ref{fig:intro_2}, \sysname accurately localizes target instances for object-goal, demand-driven, and language-guided navigation tasks. Notably, the heatmaps reveal that even non-top candidates retain task-relevant attributes. For example, in the object-goal case (\eg, “a pair of lounge chairs”), secondary matches preserve key spatial and functional cues such as seating layout and back-to-back configuration. In demand-driven scenarios (\eg, “Where should I eat?”), nearby tables also reflect contextual relevance.

\subsection{Ablation Studies}
\label{sec:exp_ablation}

\noindent \textbf{Ablation Study on Mapping.}  
In \tab\ref{table:ablation_map}, we evaluate the impact of two key components in \sysname's mapping pipeline (\S\ref{sec:design1}): structural consensus (Structural.) and semantic similarity (Semantic.). When using only structural consensus, the AP drops from 14.3 to 12.2, yet remains higher than all zero-shot baselines.
Moreover, since spatial structural relations are essential for associating instances in 3D reconstruction, structural constraints cannot be entirely removed. To evaluate the performance of using only semantic similarity, we adopt the local structural similarity metric from OVIR-3D as a baseline. The result shows a significant AP drop of 4\%.
When both components are used jointly, \sysname achieves the best performance, as expected. These results confirm that both structural and semantic cues are critical to the effectiveness of \sysnameposs mapping strategy.

\noindent \textbf{Ablation Study on Instruction Grounding.}  
\tab\ref{table:ablation_nav} presents an ablation study on two key strategies in the instruction-to-instance grounding (\S\ref{sec:design2}): unconstrained instruction parsing without restricting LLM outputs to a predefined vocabulary (Ins. Parsing), and \sysname-assisted instance selection (Ins. Selec.). When Ins. Parsing is removed, we follow the parsing approach used in VLMap as a baseline.
Experimental results show that removing both strategies leads to a drop of over 10\% in SR compared to the full \sysname pipeline. Nevertheless, due to the accurate semantic map, our method still outperforms VLMap and NLMap (see \tab\ref{table:nav}), and achieves better performance than ConceptGraphs on SR$_8$ and SR$_{16}$.
Individually, removing Ins. Parsing results in a 4.9\% decrease in SR, while excluding Ins. Selec. causes a 2.4\% drop. Notably, since Ins. Selec. performs filtering within the top-8 most relevant candidates, its removal does not affect SR$_8$ and SR$_{16}$.
The results demonstrate that \sysname effectively unlocks the instruction interpretation capabilities of LLMs.

\noindent \textbf{Ablation Study on Hyperparameters.}  
We conducted additional evaluations to assess the robustness of our algorithm with respect to key hyperparameters. As shown in \tab\ref{table:ablation_p1}, we first examined the effect of the consensus threshold used in the mapping stage. Within the range of 0.5–0.7, the AP variation remains within 0.34. The lowest performance occurs at a threshold of 0.7, yielding an AP of 13.7, while the highest AP of 14.3 is achieved at 0.6, which we adopt in practice.
We further evaluated the impact of the number of candidate instances used in instruction-to-instance grounding. As shown in \tab\ref{table:ablation_p2}, SR remains stable within a 1.9\% fluctuation when the candidate count ranges from 4 to 12, with the best performance (49.6\% SR) observed at 8 candidates.
These ablation results demonstrate the consistent and robust performance of \sysname across a range of hyperparameter settings.

\begin{table}[t]
\centering
\begin{tabular}{l|ccc}
\toprule
 & AP & AP$_{50}$ & AP$_{25}$ \\
\midrule
$\tau_{thres}$ (0.5 -- 0.7) & $14.0 \pm 0.34$ & $36.9 \pm 1.14$ & $49.4 \pm 2.41$ \\
\bottomrule
\end{tabular}
\caption{Impact of consensus threshold.}
\label{table:ablation_p1}
\end{table}

\begin{table}[t]
\centering
\begin{tabular}{l|ccc}
\toprule
 & SR[\%] & SR$_{8}$[\%] & SR$_{16}$[\%] \\
\midrule
$N_{c}$ (4-12) & $47.7 \pm 1.9$ & $57.6 \pm 0.7$ & $68.2 \pm 0.0$ \\
\bottomrule
\end{tabular}
\caption{Impact of candidate number.}
\label{table:ablation_p2}
\end{table}

\section{Conclusion}
\label{sec:conclusion}

We present OpenMap, a zero-shot open-vocabulary visual-language mapping framework for accurate instruction grounding in embodied navigation. To address challenges in instance inconsistency and limited instruction expressiveness, we propose a structural-semantic consensus constraint for robust 3D instance aggregation and an instruction-to-instance grounding module for fine-grained grounding of free-form commands. Extensive experiments on ScanNet200 and Matterport3D demonstrate that \sysname consistently improves both semantic mapping and instruction to target instance retrieval under zero-shot settings. By enabling precise alignment between natural language instructions and 3D scene instances, \sysname makes a concrete step toward more reliable and generalizable instruction execution in real-world navigation tasks.

\begin{acks}
We sincerely thank the MobiSense group and the Tsinghua University - Inspur Yunzhou Joint Research  Center for New Industrialization and Trustworthy Networks. This work is supported in part by the National Key Research Plan under grant No. 2021YFB2900100, the NSFC under grant No. 62372265, No. 62302254, and No. 62402276.
\end{acks}

\bibliographystyle{ACM-Reference-Format}
\bibliography{reference}

\end{document}